\documentclass{article}

\usepackage{PRIMEarxiv}

\usepackage[utf8]{inputenc} 
\usepackage[T1]{fontenc}    
\usepackage{hyperref}       
\usepackage{url}            
\usepackage{booktabs}       
\usepackage{amsfonts}       
\usepackage{nicefrac}       
\usepackage{microtype}      
\usepackage{lipsum}
\usepackage{fancyhdr}       
\usepackage{graphicx}       

\usepackage{graphics} 
\usepackage{epstopdf} 
\usepackage{epsfig} 
\usepackage{xcolor}
\usepackage{subcaption}
\usepackage{lipsum}
\usepackage{mathptmx} 
\usepackage{times} 
\usepackage{amsmath} 
\usepackage{amssymb}  
\usepackage{filecontents}
\usepackage{cite}
\usepackage{wasysym}
\usepackage{bm}
\usepackage{multirow}
\graphicspath{{media/}}     

\title{Emergency-braking Distance Prediction using Deep Learning

}

\author{
  Ruisi~Zhang\\
  Department of Mechanical Engineering \\
  University of Utah \\
  Salt Lake City\\
  \texttt{ruisi.zhang@utah.edu} \\
   \And
  Ashkan~Pourkand \\
  Department of Computer Science \\
  University of Utah \\
  Salt Lake City\\
  \texttt{ashkan.pourkand@utah.edu} \\
}

\begin{document}
\maketitle

\begin{abstract}
Predicting emergency-braking distance is important for the collision avoidance related features, which are the most essential and popular safety features for vehicles. In this study, we first gathered a large data set including a three-dimensional acceleration data and the corresponding emergency-braking distance. Using this data set, we propose a deep-learning model to predict emergency-braking distance, which only requires 0.25 seconds three-dimensional vehicle acceleration data before the break as input. We consider two road surfaces, our deep learning approach is robust to both road surfaces and have accuracy within 3 feet. 
\end{abstract}


\section{Introduction}
Safety of autonomous vehicles has been one of the main concerns for decades, especially with the rapid growth of self-driving technology recently, which motivates researchers to develop advanced safety features for vehicles. The collision avoidance related features are the most essential and popular safety features. The commercially available features including forward-collision warning or avoidance (FCW/FCA) and autonomous emergency braking (AEB) has significantly reduced the front-to-rear crash rates in recent years \cite{cicchino2017effectiveness}, which are offered as an optional feature set for more than half of U.S. vehicle series. 

The existing FCW or AEB methods utilized the time-to-collision (TTC) \cite{lee1976theory,van1993time,shaw1996vehicle,miller2002adaptive,yang2003development,yang2004vehicle} or the minimum allowable distance to stop before the collision \cite{katsumata1978radar,kiefer1999development,kiefer2003forward} as the major cue for decision-making in traffic. Yang et al. \cite{yang2003development} suggested a method for FCW that triggers the warning when TTC, which determined by the relative distance and velocity to the target, is less than a 2 seconds threshold. Dagan et al. \cite{dagan2004forward} improved the method using TTC threshold by considering the additional effect of relative accelerations to the target, which is particularly useful for the scenarios of a sudden stop or slowing down of the target vehicle. 

TTC can accurately characterize human factors including drivers reaction time \cite{lange2017data}, braking type \cite{delaigue2004comprehensive,he2018regenerative}, etc. However, human factors can be neglected for AEB and self-driving technologies, which is the current trend, since braking is not controlled by drivers but the vehicle itself. This led us to focus on the minimum distance needed to stop before the collision. Katsumate et al. \cite{katsumata1978radar} initially determined the minimum allowable distance by a function of two input variables: current speed and the angle of steering movement. This function considered the factors of the road and weather conditions as a constant effect, which failed to characterize the complexity of dynamic driving conditions. Kiefer \cite{kiefer1999development,kiefer2003forward} then proposed an improved linear function using dynamic parameters to predict the minimum allowable distance. Wang \cite{wang2016development} later applied an advanced driving simulator to further improve the algorithm for a wide range of kinematic conditions. 

\begin{figure*}[t!]
	\centering
	\begin{subfigure}[b]{0.255\textwidth}
		\centering		
		\includegraphics[width=\columnwidth]{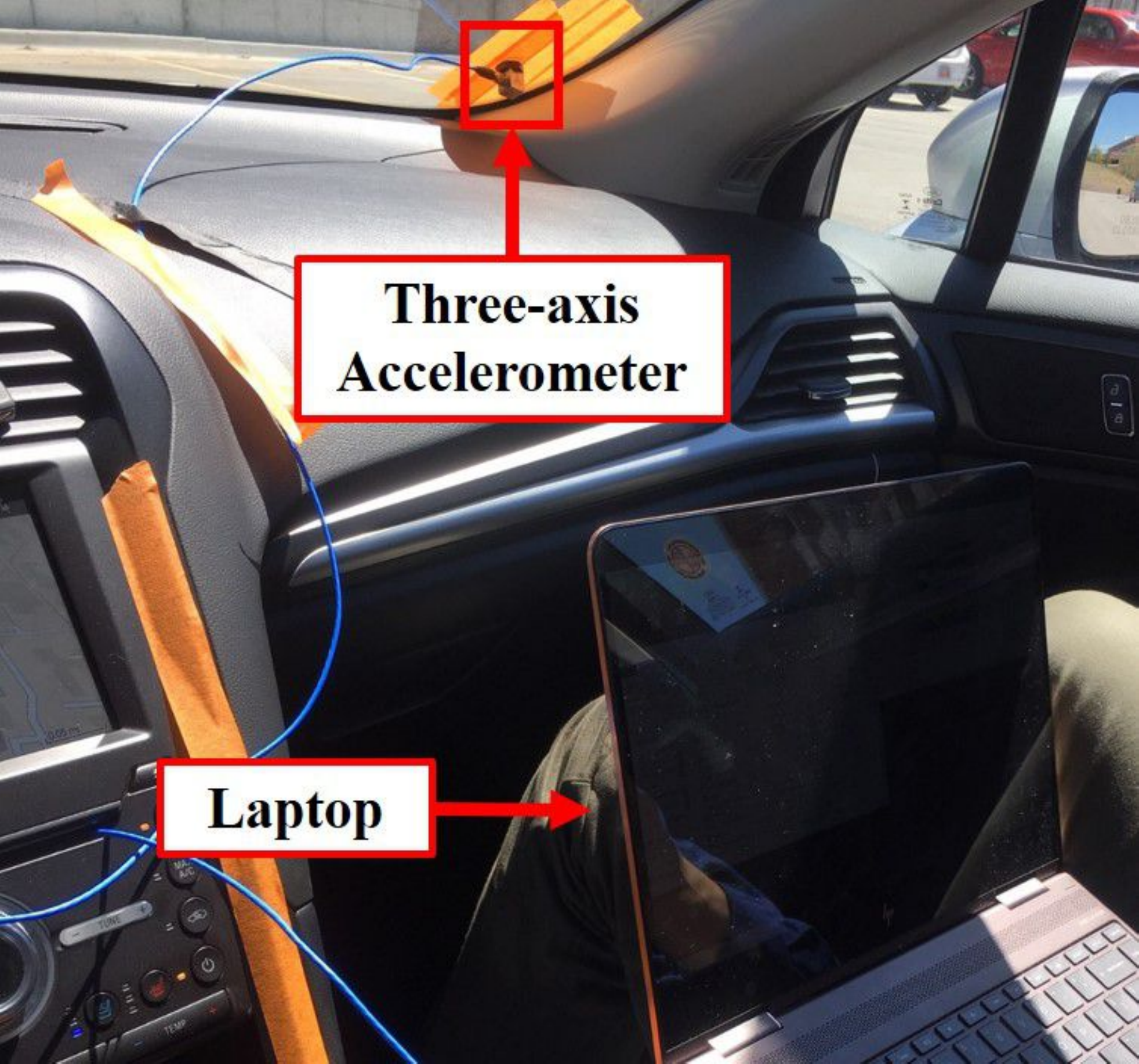}
		\caption{}
		\label{inside}
	\end{subfigure}
	\hfill
	\begin{subfigure}[b]{0.3708\textwidth}
		\centering
		\includegraphics[width=\columnwidth]{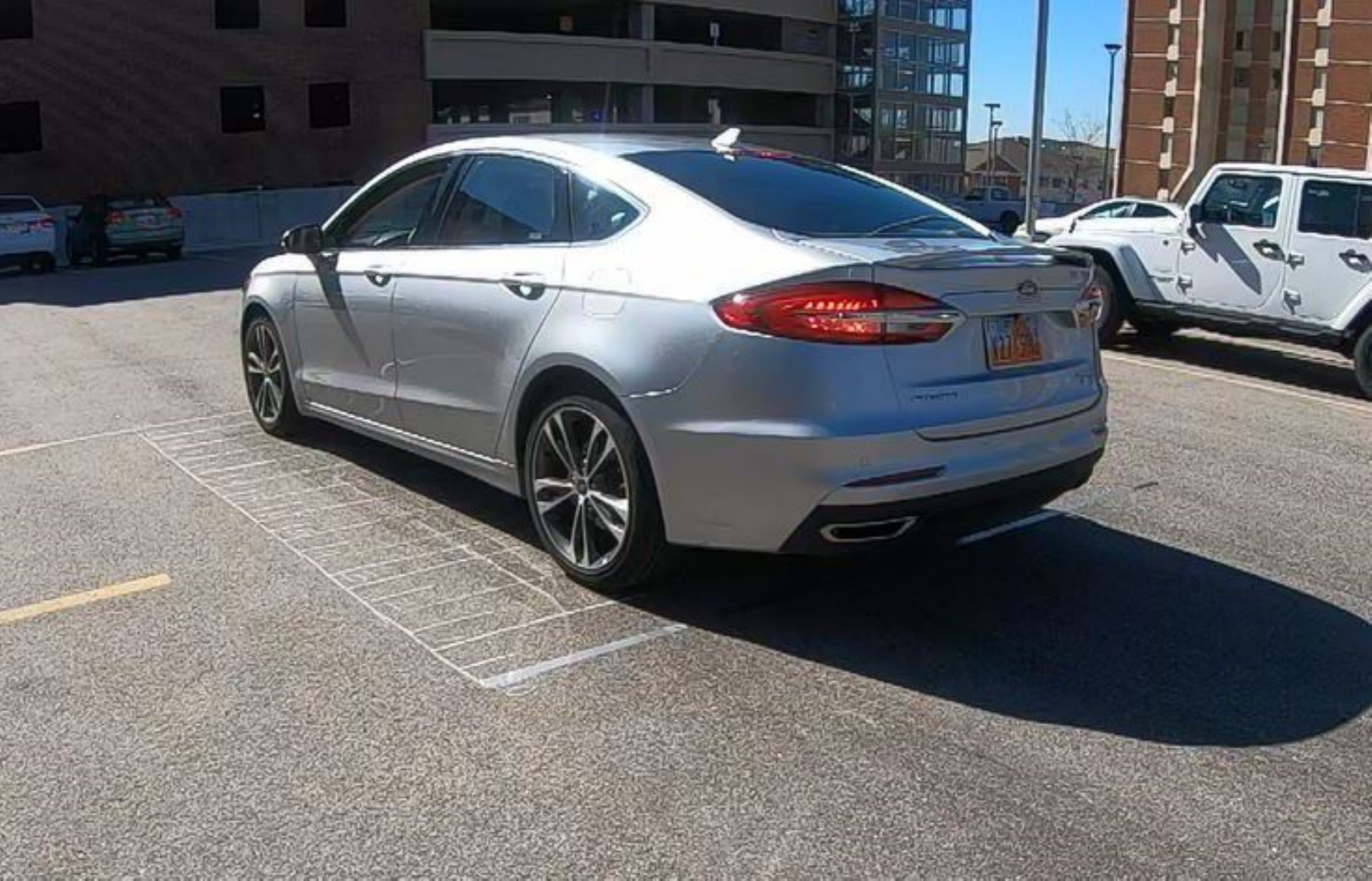}
		\caption{}
		\label{outside:day}
	\end{subfigure}	
	\hfill
	\begin{subfigure}[b]{0.352\textwidth}
		\centering
		\includegraphics[width=\columnwidth]{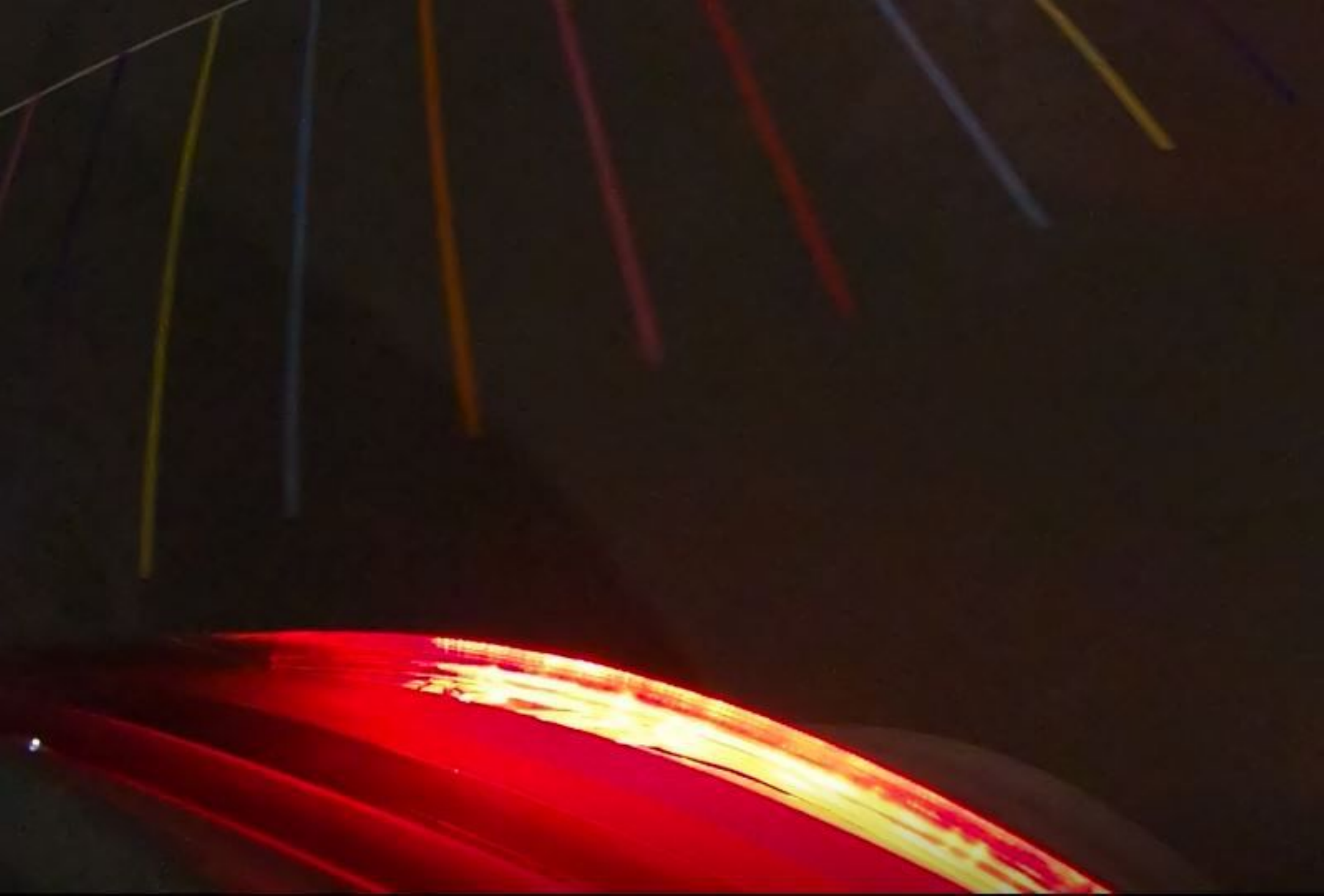}
		\caption{}
		\label{outside:night}
	\end{subfigure}	
	
	\caption{Data collection setup. The entire emergency-braking process is recorded by a stationary camera and a three-axis accelerometer. The braking distance is measured using the floor markers in 0.30\,m (1\,ft) increments from when the brake light is engaged to when the vehicle fully is stopped. (a) Inside, a accelerometer is rigidly attached to the front windshield of the vehicle and connected to a laptop. (b) Outside, a camera is placed on the tripod pointing to the entire floor markers with 45 degrees angle for daytime use. (c) Outside, a camera is rigidly attached using a suction cup mount to the back of the car pointing to the brake light and floor for nighttime use.
	}
	\label{fig:setup}	
\end{figure*}

Although these early proposed approaches with linear models were widely used in the automotive industry, linear models are not ideal for the complex dynamic driving conditions even with dynamic inputs \cite{katare2019embedded}. The most recent machine learning-based approaches for the FCW system showed better performance in characterizing the non-linearities of the complex driving environments by using non-linear active functions to extract abstract patterns or features from the inputs. Katare and El-sharkawy \cite{katare2019embedded} using a fully connected neural network with two hidden layers to classify whether or not triggering the warning, which trained by radar-based inputs, vehicle velocity, acceleration and the separation distance from the front objects, and ground truth labels. However, machine learning-based approaches required preprocessed data, since raw data might be too complicated machine learning algorithms, like decision trees, neural networks, etc., and their performances relied on both encoding and the algorithms' structure. We proposed a deep learning-based approach, which can directly input raw data without any data encoding or editing, to predict the minimum allowable distance defined as the distance between the position of the vehicle when emergence-brake starts and the position that the vehicle is completely stopped. 


One challenge for modern FCW and AEB systems is the reliability of certain weather conditions. Almost all existing FCW and AEB algorithms are heavily dependent on the sensors (radar, lidar, or camera), which might be blocked by ice, snow, or glare at sunrise or sunset. With the development of wireless communication technology, the cooperative vehicle collision warning safety feature that utilized peer-to-peer data sharing via GPS information could be one solution for this problem \cite{miller2002adaptive, flores2018cooperative}. However, this feature highly relied on the vehicle-to-vehicle communication system that is not currently commercially available. We utilize a three-dimensional acceleration data measured by a three-axis accelerometer attached to the windshield inside of the vehicle as the only input, which will not be affected by any possible extreme weather conditions. 

The other challenge is the robustness on various road surfaces and conditions since most FCW and AEB algorithms are designed for a single road surface under optimal conditions. Previous studies \cite{giguere_simple_2011,pourkand,Graham,Macleod} proposed various surface classification methods for optimizing the performance of the vehicle based on the surface. Giguere and Dudek \cite{giguere_simple_2011} designed a simple probe to identify the surface consisting of an accelerometer at the end of a bar that was dragged behind the vehicle while traveling at low speed. Graham and Sutcliffe \cite{Graham} studied the classification of asphalt using a roller and images to grade and classify the asphalt. DuPont et al. \cite{Dupont} utilized similar methods with a low-frequency accelerometer placed on the body of the unmanned ground vehicle (UGV) systems. Macleod et al. \cite{Macleod} developed a robot to conduct a nondestructive evaluation of the roads based on the vibration using a neural network. Ojeda et al. \cite{ojeda06} instrumented a Pioneer 2-AT robot with a variety of sensors to track robot linear and angular acceleration (accelerometer and gyroscope), wheel movement (encoders) and motor effort, and road-robot interaction (IR, ultrasonic range finder, and microphone). They processed the sensor outputs using both time- and frequency-based features. The resulting feature vectors and known surface classes were then used to train a two-layer back-propagation network. They found that the single-axis data of the accelerometer overall outperformed other inputs for surface classification with exceptions of grass was best identified by the microphone and sand was by the measured motor effort. However, none of the studies were designed to be directly applicable to vehicles due to the lack of vehicle suspension or placement of the accelerometer. 

Instead of road surfaces classification, we proposed a deep learning approach to predict the emergency-braking distance on two of the most common road surfaces, asphalt and concrete.

\section{Data Collection}
\subsection{Apparatus}
We captured the emergency-braking data using a Ford Fusion in the model year 2018, shown in Fig.\,\ref{fig:setup}. The driving and braking condition is optimal, which has been checked before processing the tests. The data collection test is performed by the authors, each of whom has a valid Utah driver's license. The study was approved by the University of Utah Department of Public Safety.

A triaxial ICP accelerometer (356A02, PCB Peizotronics) rigidly attached on the front windshield measures the vehicle's vibration at 25\,kHz. A USB analog-to-digital DAQ device from National Instruments records the three-dimensional acceleration data using a NI9264 module with NI LabVIEW software. To avoid confounding environmental factors, we turn off the climate control and stereo in this study. 

Numerous camera has been considered to measure the emergency-braking process, but none fit the required size, resolution, and mounting specification. Consequently, a GoPro HERO7 Black is chosen, capable of being controlled remotely, shooting frames at the rate of 240 frames per second in full HD, and mounting on a car at high speeds with a commercially available industrial-strength suction cup mount. For stationary use, we utilize a flex clamp combined with a Sunpak PlatinumPlus 5858D tripod to hold the camera in place.

\begin{figure*}[t!]
	\centering
	\begin{subfigure}[b]{0.3\columnwidth}
		\centering		
		\includegraphics[width=\columnwidth]{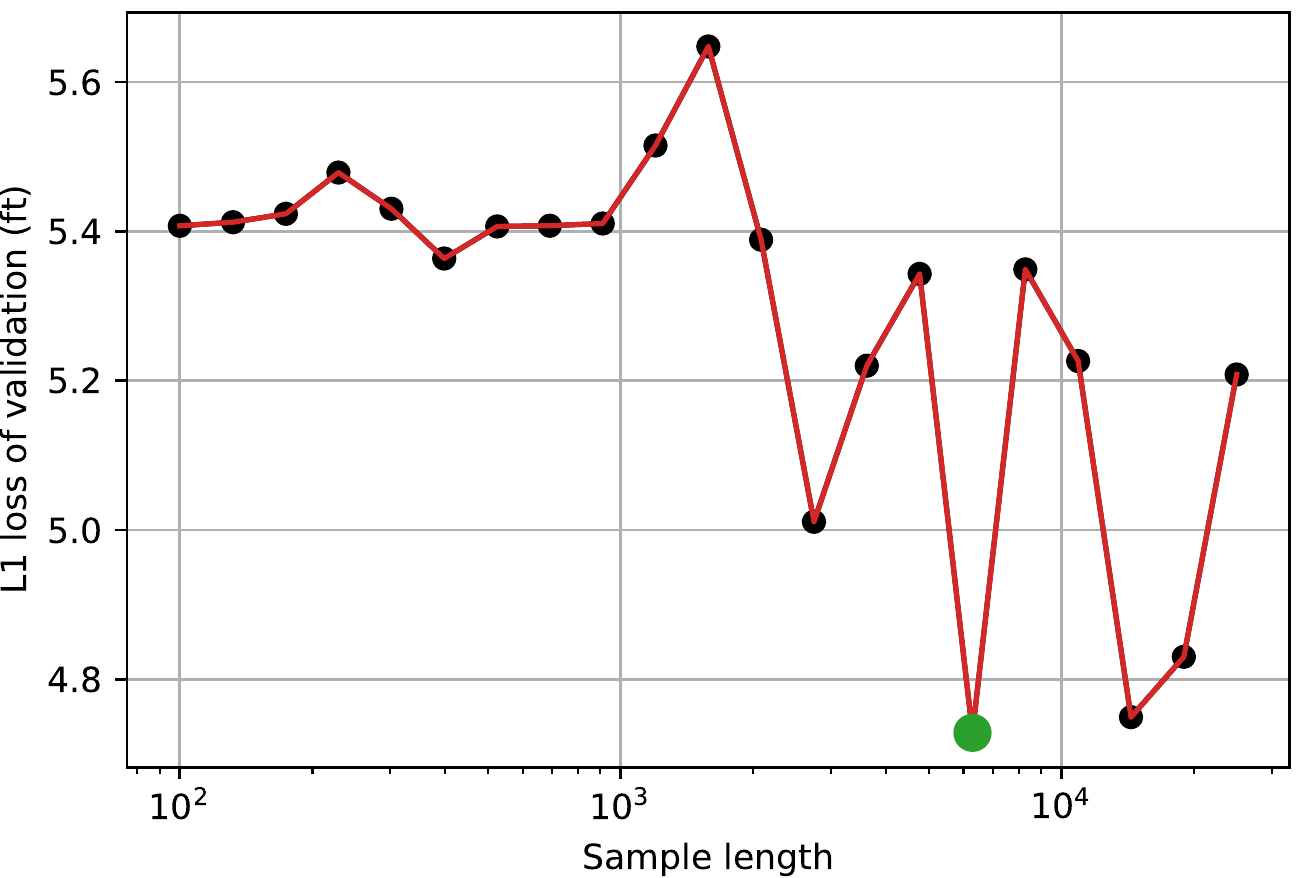}
		\caption{}
		\label{fig:mse_length}
	\end{subfigure}
	\hfill
	\begin{subfigure}[b]{0.3\columnwidth}
		\centering
		\includegraphics[width=\columnwidth]{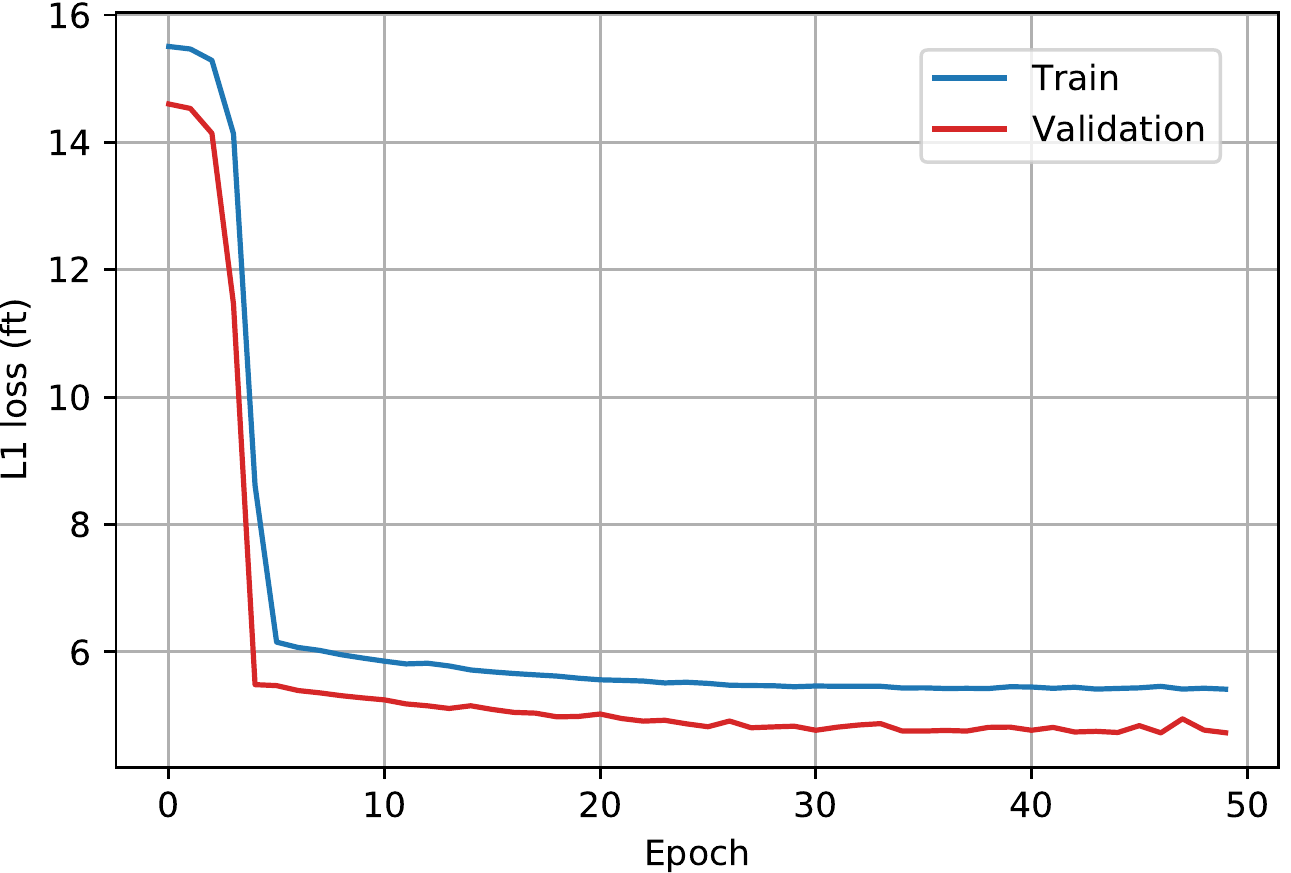}
		\caption{}
		\label{fig:L6288}
	\end{subfigure}	
	\hfill
	\vspace{1mm}
	\begin{subfigure}[b]{0.385\columnwidth}
		\centering
		\includegraphics[width=\columnwidth]{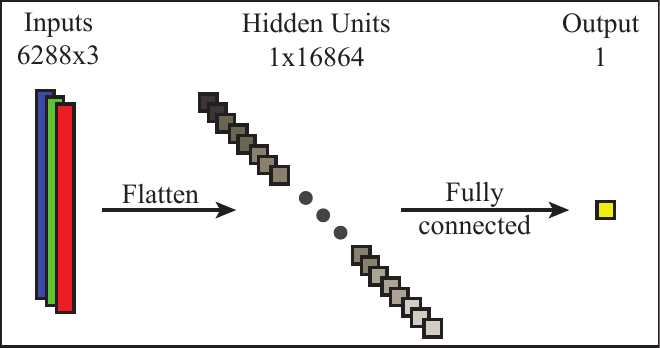}
		\caption{}
		\label{fig:baseline_network}
	\end{subfigure}	
	
	\caption{(a) The L1 losses for input with sample length in the range of 100--25,000. We chose the sample length of 6288 with the lowest loss occurs, shown as the green marker, to be the optimal input length. (b) The L1 losses for the chosen sample length for 50 epochs. (c) The network architecture for the chosen sample length. The input acceleration data has been flatten to a one-dimensional array for a linear fully connected layer to output the predicted braking distance.}
	\label{fig:sample_size}	
\end{figure*}

\begin{figure}[t]
	\centering
	\includegraphics[width=0.5\columnwidth]{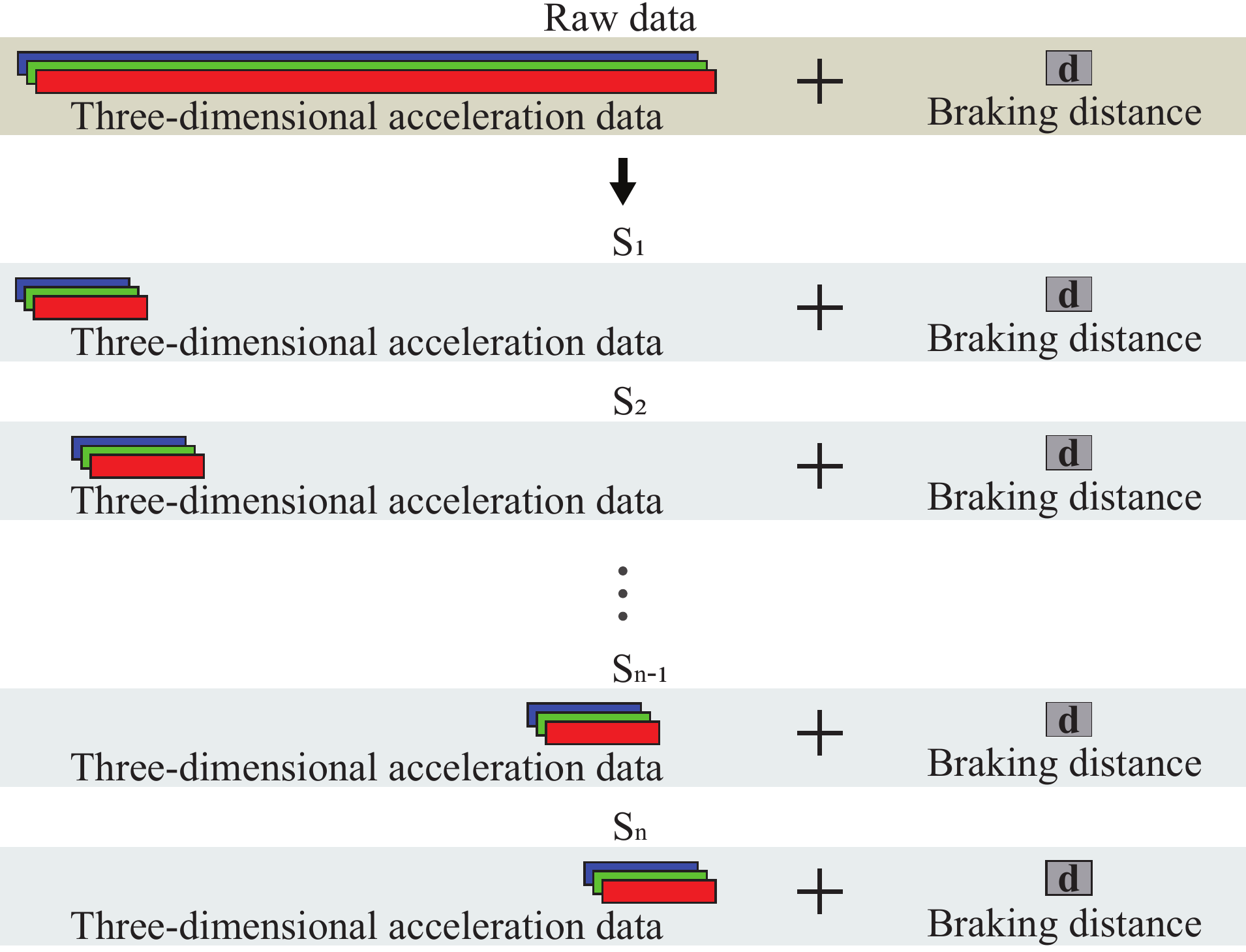}
	\caption{Acceleration data segmentation for a session. The three-dimensional acceleration data for each session has been trimmed into equal length sub-datasets with 50\% overlap between neighbors 
		with the same braking distance label.}
	\label{fig:data_preparation}	
\end{figure}
\subsection{Procedure}
In this study, we tested two of the most common road surfaces: asphalt and concrete. The data collection test is conducted in 50 sessions (25 sessions per surface), with each session containing a driving period with a constant speed followed by an emergency-brake and lasting 8 - 15 seconds. The first 25 sessions are conducted for asphalt, and the last 25 sessions are conducted for concrete. The sessions are separated by at least 5 minutes to cool down the brake pads. To mitigate the effect of the overheated brake and driver fatigue, after every five sessions, the driver is forced to take a break, and the vehicle is powered off for at least 20 minutes. 

For each surface, three constant speeds with 0.89\,{m/s} (2\,mph) error considered for the driving period of each session are 6.71\,{m/s} (15\,mph), 8.94\,{m/s} (20\,mph) and 11.2\,{m/s} (25\,mph).  we consider 10 sessions for each of the two lower speeds and 5 sessions for the highest speed due to safety concerns. 

For every session, a vehicle is driven with a constant speed in a straight path for at least 1.5\,s followed by an emergency-brake when the rear wheel is in the zone with the floor markers. This entire session is recorded by a camera, which is located outside of the vehicle for daytime use and rigidly attached to the back of the vehicle for nighttime use, and a three-axis accelerometer, which is rigidly attached to the front windshield, as shown in Fig.\,\ref{fig:setup}. We utilize the recorded video to manually measure the braking distance from when the brake light is engaged to when the vehicle is stopped. The recorded acceleration data associated with time for each session has been saved as individual CSV file for each session. Since we only consider the acceleration data for the constant speed driving period, we manually mark the starting time and ending time for this period.

\begin{figure*}[t!]
	\centering
	\begin{subfigure}[b]{0.85\columnwidth}
		\centering
		\includegraphics[width=\columnwidth]{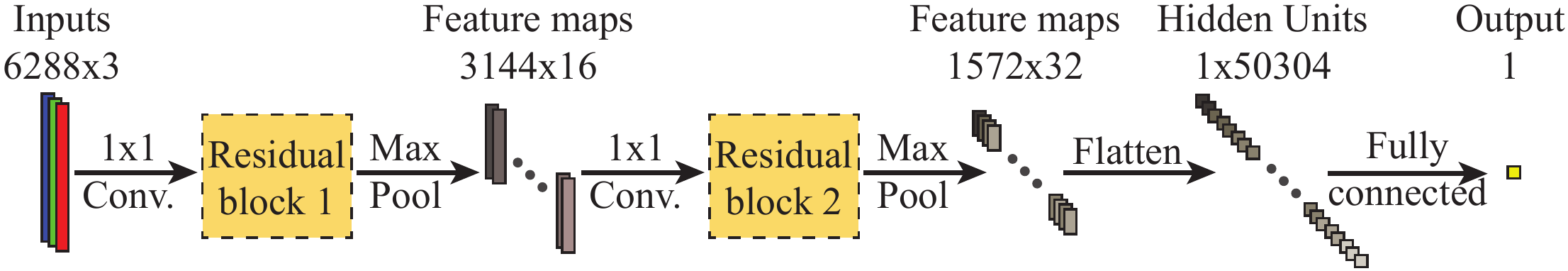}
		\caption{}
		\label{fig:CNN}
	\end{subfigure}	
	\vfill
	\vspace{1mm}
	\begin{subfigure}[b]{0.85\columnwidth}
		\centering
		\includegraphics[width=\columnwidth]{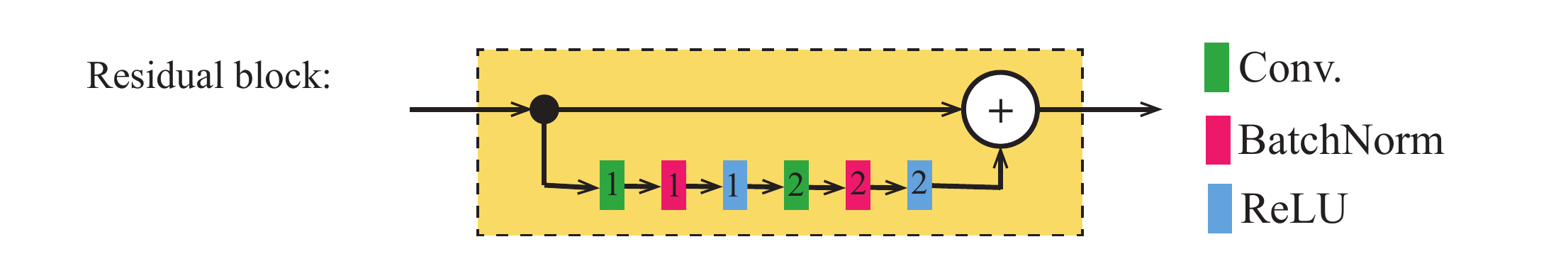}
		\caption{}
		\label{fig:res_block}
	\end{subfigure}	
	
	\caption{The proposed architecture for input with the chosen size of 6288x3. The CNN learns to produce the coefficients for extracting the features. The final fully connected layer outputs the predicted braking distance in foot.}
	\label{fig:network}	
\end{figure*}

\begin{table}[t!]
	\centering
	\begin{tabular}{|c|c|c|c|}
		\hline
		&              & Residual block 1 & Residual block 2 \\ \hline
		\multirow{4}{*}{Conv. 1} & Kernel       & 15               & 7                \\ \cline{2-4} 
		& Padding      & 7                & 3                \\ \cline{2-4} 
		& in\_channel  & 16               & 32               \\ \cline{2-4} 
		& out\_channel & 16               & 32               \\ \hline
		\multirow{4}{*}{Conv. 2} & Kernel       & 7                & 3                \\ \cline{2-4} 
		& Padding      & 3                & 1                \\ \cline{2-4} 
		& in\_channel  & 16               & 32               \\ \cline{2-4} 
		& out\_channel & 16               & 32               \\ \hline
	\end{tabular}
	\caption{Parameters for the two residual blocks for both the network with one-dimensional inputs and the one with two-dimensional inputs. The filter size for two-dimensional inputs is also two-dimensional with equal width and hight.}
	\label{tb:parameters}
\end{table}

\section{Data Preparation}
We obtained at least 1.5\,s three-dimensional acceleration data at the sampling rate of 25\,kHz, which yields at least 125,000 sample points, for each session. We trimmed the acceleration data into equal length sub-datasets, which are labeled with the same braking distance, as shown in Fig.\,\ref{fig:data_preparation}. The trimmed three-dimensional acceleration data are considered as a three-channel one-dimensional array with a given sample length. 
The 21 sample lengths for each sub-dataset considered in this study are spaced evenly in a base-10 logarithmic scale within the frequency range of 100--25,000. We shuffled all the trimmed data and split into three data loader for training, validation, and testing. We used a simple linear network with a fully-connected layer, shown in Fig.\,\ref{fig:baseline_network}, to find the optimal sample length. This network is also employed as our baseline for later analysis. We chose L1 loss as the loss function, which takes the mean element-wise absolute value difference, since it has better overall performance and more intuitive. The L1 loss means the average difference between the predicted braking distance and the ground truth with the unit of foot. 

We chose the sample length of 6288 with the lowest loss occurs, shown as the green marker in Fig\,\ref{fig:mse_length}, to be the optimal input length. Using the accelerometer at the sampling rate of 25\.kHz, the sample with the chosen length is approximately 0.25\,s acceleration data recording. We conduct two network in Sec.\,\ref{sec:network} for each of the input formats: the three-channel one-dimensional vector and the three-channel two-dimensional matrix. 

The trimmed three-dimensional acceleration data are considered as three-channel one-dimensional vector, which are then reshaped as three-channel two-dimensional matrix by stacking data row by row. Since the square root of optimal input data length  of 6288 is not integer, we decided to use the input data length of 10,000 for two-dimensional input data for convenience.  This input data containing 10,000 sample points yields a three-channel 100x100 two-dimensional acceleration matrix. 

We shuffle all the trimmed data and split into three data loader for training, validation, and testing. We conduct two network in Sec.\,\ref{sec:network} for each of the input formats: the three-channel one-dimensional vector and the three-channel two-dimensional matrix.

\section{Deep-Learning Approach}

\subsection{Network Representation} \label{sec:network}

We use two convolutional neural network (CNN) followed by a fully connected layer with the same architecture for both the three-channel one-dimensional acceleration input data, as shown in Fig\,\ref{fig:network}, and the three-channel two-dimensional acceleration input data. The only differences for one-dimensional inputs and two-dimensional inputs are the dimensions of filters. The length of all the filters used in two network are identical, the one for two-dimensional network has the filters with both width and height equal to the length of the one for one-dimensional network. the We employed two residual blocks to extract the features from the input acceleration data associated with a convolutional layer located before and a max pooling layer located after. The convolutional layer used before residual block is using a 1x1 kernel size filter that will reshape the data to match the data size of the output of the residual block for summation. The max pooling layer has kernel size of 3 and stride of 2, to extract the feature information. Each residual block has two convolutional layers followed by a batch normalization and a non-linear activation function of ReLU, an elementwise activation function eliminates the negative values. The detailed values for each parameter is shown in Table\,\ref{tb:parameters}. We utilize L1 loss function as same as the baseline network and stochastic gradient descent as out optimizer with learning rate of 5e-6 and momentum of 0.9.


\subsection{Results}
\begin{figure}[t!]
	\centering
	\includegraphics[width=0.5\columnwidth]{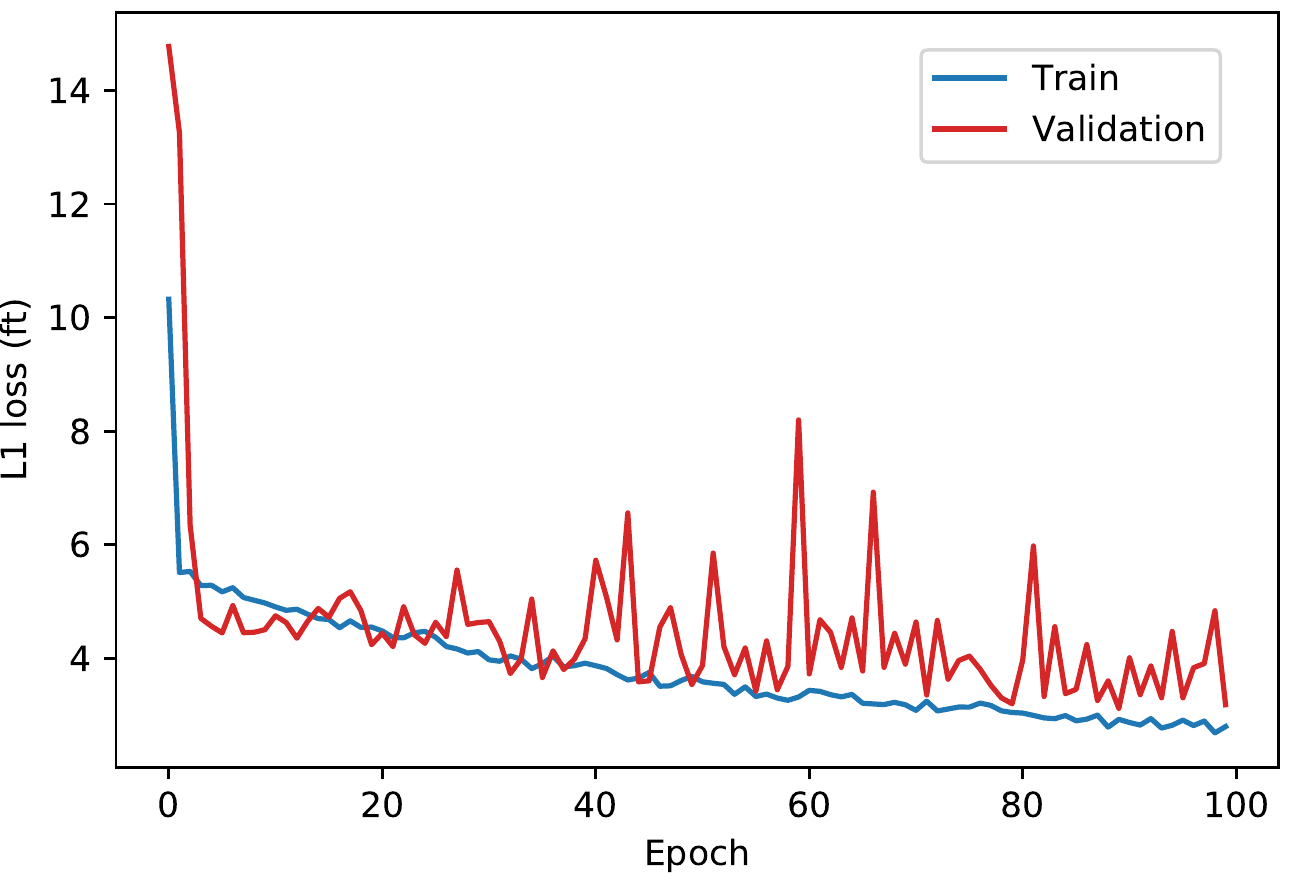}
	\caption{The L1 losses for the proposed one-dimensional deep learning approach. The L1 loss of the best model on test dataset is 3.79\,ft.}
	\label{fig:results}	
\end{figure}

\begin{figure}[t!]
	\centering
	\includegraphics[width=0.5\columnwidth]{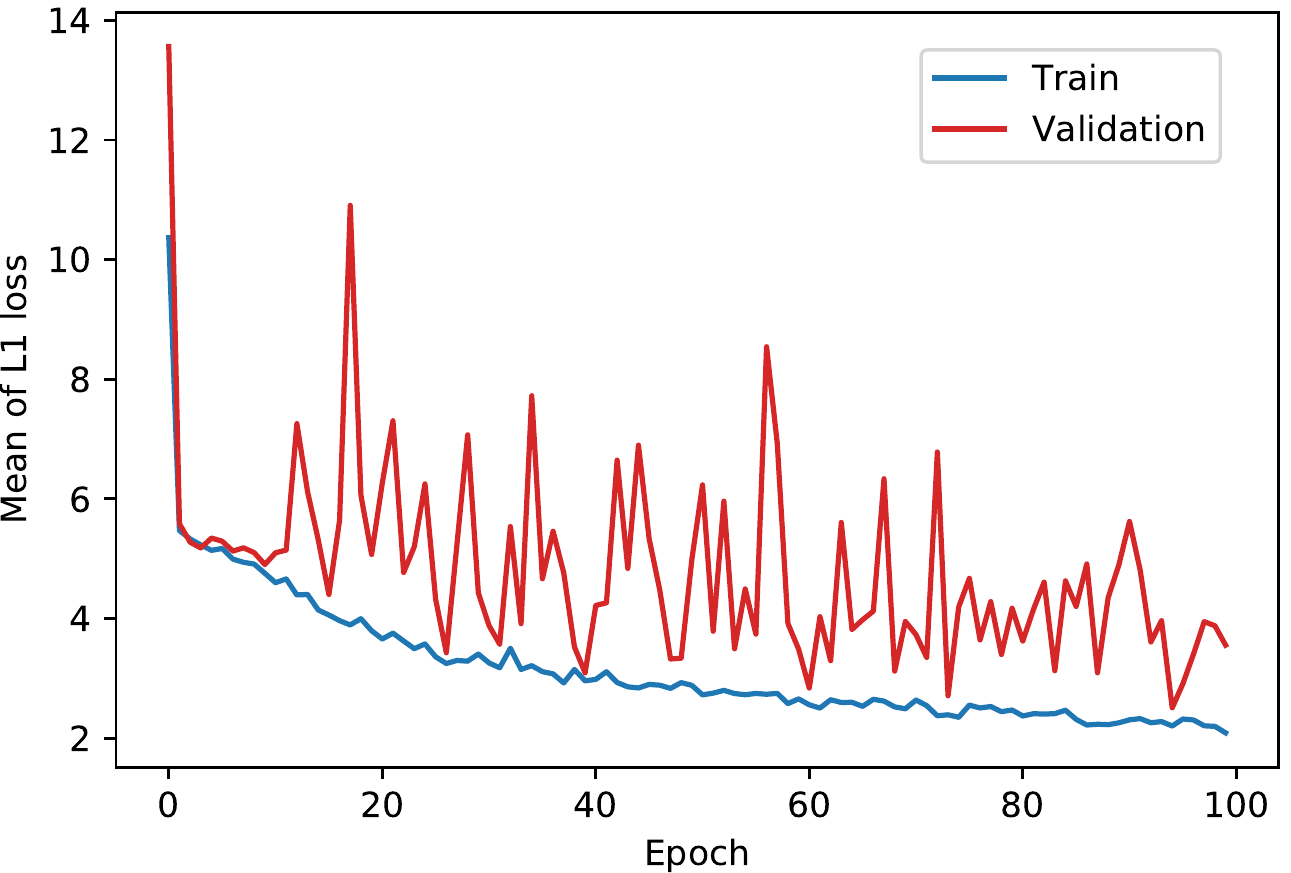}
	\caption{The L1 losses for the proposed two-dimensional deep learning approach. The L1 loss of the best model on test dataset is 3.19\,ft.}
	\label{fig:results_2d}	
\end{figure}

Fig.\,\ref{fig:results} shows the L1 losses for both training and validation datasets, using one-dimensional input. While the training loss was steadily decreasing, the validation set was noisy and did not improve significantly after the 30th epoch. Interestingly, the network performed better on the validation set than the training set. Fig.\,\ref{fig:results} shows the losses for both training and validation datasets, using the same CNN network but with two-dimensional input.  In this case, the networked trained faster on the training set and the loss decreased faster than the one-dimensional input, but training loss did not improve and the networked overfitted on the training dataset. After 100 epochs, the best model for one-dimensional input has L1 loss of 3.79\,ft on the test dataset and the best model for two-dimensional input has L1 loss of 3.19\,ft on the test dataset.

\section{Discussion}
Comparing with the results we obtained from the simple linear neural network with one fully connected layer which is the chosen baseline, both one-dimensional CNN and two-dimensional CNN showed better performance with 1 foot less on the mean of L1 loss for both validation dataset and testing dataset. Although the input data contains more sampling points for two-dimensional CNN, the performance is not significant better than the one of one-dimensional CNN. We suggest to directly use one-dimensional acceleration data as input and train the one-dimensional network. 

We tested various filter size for all convolutional layers. We found that relatively small filter size for the first residual block did not perform well, meaning the mean of L1 loss less than 4. Consequently, we chose relatively large filter size for the convlolutional layer. We found that relatively small filter for the second residual block performed better. We suggest the convolutional layers in the first residual block are attempt to learn the coarse features and those in the second residual block are attempt to learn the fine features. In addition, the input size for these two residual blocks is not the same, since the first one is before the max pooling layer and the second one is after.

Before using the residual blocks, we directly connect 4 convolutional layer in series, which did not provide better results than the baseline. We suggest residual blocks works better for this task. 

For the two tested road surface (i.e., asphalt and concrete), we highly recommended to utilize our model to predict the emergency-braking distance since our model is robust to various road surfaces and have accuracy with approximately 3 feet, which will be useful for FCW and AEB. The robustness of our model suggests our model are capable of capturing the effect of dynamic driving conditions. Although we only considered two road surface, for future work, we suggest to collect more data using our proposed data gathering procedure for other surfaces to test the robustness of this model.

\section{Conclusion}
In this study, we first gathered a large data set including a three-dimensional acceleration data and the corresponding emergency-braking distance. Using this data set, we propose a deep-learning model to predict emergency-braking distance, which only requires 0.25 seconds three-dimensional vehicle acceleration data before the break as input. We consider two road surfaces, our deep learning approach is robust to both road surfaces and have accuracy within 3 feet.  


\appendix  
We saved all the raw acceleration data for each trail as CSV file under the zip folder named {'brake\_{data}\_csv.csv'}. We saved the label, CSV file name, valid start index, valid stop index, and speed for each trail as the other CSV file named 'data\_entry.csv'. 

We use PyTorch 1.0 as our deep learning framework. We used CADE machine to train our model. We create our own class for training, validation, and testing datasets. Based on these datasets, we create dataloader. We wrote our code using Jupyter, the Jupyter file for one-dimensional input is named 'project\_1d.py' and the Jupyter file for two-dimensional input is named 'project\_2d.py'.


%



\section*{Acknowledgment}

The authors would like to thank Mohammad Homayounpour , Tolga Tasdizen, and Ricardo Bigolin Lanfredi for their advice on designing the architecture of network and reporting the results.

\bibliographystyle{unsrt}  
\bibliography{references}

\begin{thebibliography}{10}

\bibitem{cicchino2017effectiveness}
Jessica~B Cicchino.
\newblock Effectiveness of forward collision warning and autonomous emergency
  braking systems in reducing front-to-rear crash rates.
\newblock {\em Accident Anal. Prevention}, 99:142--152, 2017.

\bibitem{lee1976theory}
David~N Lee.
\newblock A theory of visual control of braking based on information about
  time-to-collision.
\newblock {\em Perception}, 5(4):437--459, 1976.

\bibitem{van1993time}
Richard Van Der~Horst and Jeroen Hogema.
\newblock Time-to-collision and collision avoidance systems.
\newblock In {\em Proc. 6th ICTCT Workshop—Safety Evaluation of Traffic
  Syst.: Traffic Conflicts and Other Measures}, pages 109--121, 1993.

\bibitem{shaw1996vehicle}
David~CH Shaw and Judy~ZZ Shaw.
\newblock Vehicle collision avoidance system, June~25 1996.
\newblock {US} Patent 5,529,138.

\bibitem{miller2002adaptive}
Ronald Miller and Qingfeng Huang.
\newblock An adaptive peer-to-peer collision warning system.
\newblock In {\em Proc. IEEE 55th Veh. Technol. Conf.}, volume~1, pages
  317--321, 2002.

\bibitem{yang2003development}
Lee Yang, Ji~Hyun Yang, Eric Feron, and Vishwesh Kulkarni.
\newblock Development of a performance-based approach for a rear-end collision
  warning and avoidance system for automobiles.
\newblock In {\em Proc. IEEE Intell. Vehicles Symp.}, pages 316--321, 2003.

\bibitem{yang2004vehicle}
Xue Yang, Leibo Liu, Nitin~H Vaidya, and Feng Zhao.
\newblock A vehicle-to-vehicle communication protocol for cooperative collision
  warning.
\newblock In {\em Proc. Int. Conf. MobiQuitous}, pages 114--123, 2004.

\bibitem{katsumata1978radar}
Masaaki Katsumata and Norio Fujiki.
\newblock Radar-operated collision avoidance system for roadway vehicles using
  stored information for determination of valid objects, February~7 1978.
\newblock {US} Patent 4,072,945.

\bibitem{kiefer1999development}
RJ~Kiefer, D~LeBlanc, MD~Palmer, J~Salinger, Richard~K Deering, Mike Shulman,
  et~al.
\newblock Development and validation of functional definitions and evaluation
  procedures for collision warning/avoidance systems.
\newblock Technical report, NHTSA, U.S. Dept. Transp., Washington, DC, USA, DOT
  HS 808 964, 1999.

\bibitem{kiefer2003forward}
RJ~Kiefer, MT~Cassar, CA~Flannagan, DJ~LeBlanc, MD~Palmer, RK~Deering, and
  MA~Shulman.
\newblock Forward collision warning requirements project: refining the camp
  crash alert timing approach by examining" last second" braking and lane
  change maneuvers under various kinematic conditions.
\newblock Technical report, NHTSA, U.S. Dept. Transp., Washington, DC, USA, DOT
  HS 809 574, 2003.

\bibitem{dagan2004forward}
Erez Dagan, Ofer Mano, Gideon~P Stein, and Amnon Shashua.
\newblock Forward collision warning with a single camera.
\newblock In {\em Proc. IEEE Intell. Vehicles Symp.}, pages 37--42, 2004.

\bibitem{lange2017data}
Robert Lange, Sean Kelly, Carmine Senatore, Joel Wilson, Ryan Yee, and Ryan
  Harrington.
\newblock Data requirements for post-crash analyses of collisions involving
  collision avoidance technology equipped automated and connected vehicles.
\newblock In {\em 25th Int. Tech. Conf. Enhanced Safety Vehicles Nat. Highway
  Traffic Safety Admin.}, 2017.

\bibitem{delaigue2004comprehensive}
P~Delaigue and A~Eskandarian.
\newblock A comprehensive vehicle braking model for predictions of stopping
  distances.
\newblock volume 218, pages 1409--1417, 2004.

\bibitem{he2018regenerative}
HongWen He, Bing Lu, Rui Xiong, and Jiankun Peng.
\newblock The regenerative braking control based on the prediction of braking
  intention for electric vehicles.
\newblock {\em DEStech Trans. Environment, Energy and Earth Sci.}, (iceee),
  2018.

\bibitem{wang2016development}
Xuesong Wang, Ming Chen, Meixin Zhu, and Paul Tremont.
\newblock Development of a kinematic-based forward collision warning algorithm
  using an advanced driving simulator.
\newblock {\em IEEE Trans. Intell. Transp. Syst.}, 17(9):2583--2591, 2016.

\bibitem{katare2019embedded}
Dewant Katare and Mohamed El-Sharkawy.
\newblock Embedded system enabled vehicle collision detection: An ann
  classifier.
\newblock In {\em Proc. 9th Annu. Comput. Commun. Workshop Conf.}, pages
  284--289, 2019.

\bibitem{flores2018cooperative}
Carlos Flores, Pierre Merdrignac, Raoul de~Charette, Francisco Navas, Vicente
  Milan{\'e}s, and Fawzi Nashashibi.
\newblock A cooperative car-following/emergency braking system with
  prediction-based pedestrian avoidance capabilities.
\newblock {\em IEEE Trans. Intell. Transp. Syst.}, (99):1--10, 2018.

\bibitem{giguere_simple_2011}
P.~Giguere and G.~Dudek.
\newblock A simple tactile probe for surface identification by mobile robots.
\newblock {\em IEEE Trans. Robot.}, 27(3):534--544, 2011.

\bibitem{pourkand}
Ashkan Pourkand, Chris White, Naghmeh Zamani, and David Grow.
\newblock Surface recognition for cars: A comprehensive approach for neural
  networks.
\newblock volume~3 of {\em Dynamic Systems and Control Conference}, 10 2019.

\bibitem{Graham}
W.~R. Graham, F.~Liu, M.~P.~F. Sutcliffe, and M.~Dale.
\newblock Characterisation and simulation of asphalt road surfaces.
\newblock {\em Wear}, 271(5):734--747, 2011.

\bibitem{Macleod}
Charles~Norman MacLeod, S.~G. Pierce, J.~C. Sullivan, and Anthony Pipe.
\newblock Remotely deployable autonomous surface inspection and
  characterisation using active whisker sensors.
\newblock In {\em Proc. 6th Eur. Workshop Structural Health Monitoring 1st Eur.
  Conf. Prognostics Health Manage. Soc.}, 2012.

\bibitem{Dupont}
Edmond~M. DuPont, Carl~A. Moore, Emmanuel~G. Collins, and Eric Coyle.
\newblock Frequency response method for terrain classification in autonomous
  ground vehicles.
\newblock {\em Auton. Robots}, 24(4):337--347, 2008.

\bibitem{ojeda06}
Lauro Ojeda, Johann Borenstein, Gary Witus, and Robert Karlsen.
\newblock Terrain characterization and classification with a mobile robot.
\newblock {\em J. Field Robot.}, 23(2):103--122, 2006.

\end{thebibliography}

\end{document}